\documentclass[10pt,twocolumn,letterpaper]{article}

\usepackage{iccv}
\usepackage{times}
\usepackage[pdftex]{graphicx}
\usepackage{amsmath}
\usepackage{amssymb}

\usepackage[accsupp]{axessibility}

\usepackage{caption}
\usepackage{subcaption}
\usepackage{xcolor}
\usepackage{mathtools}
\usepackage{xfrac}
\usepackage{multirow}
\usepackage{wasysym}
\usepackage{algorithm}
\usepackage{algpseudocode}

\DeclareMathOperator*{\argmax}{arg\,max}

\newcommand{\customsize}{\fontsize{9.25}{9.5}\selectfont}

\usepackage[breaklinks=true,bookmarks=false]{hyperref}

\iccvfinalcopy 

\def\iccvPaperID{1963} 

\makeatletter
\newcommand\notsotiny{\@setfontsize\notsotiny{6.5}{7}}
\makeatother

\begin{document}

\title{\vspace{-0.4cm}Fog Simulation on Real LiDAR Point Clouds \\ for 3D Object Detection in Adverse Weather}

\author{Martin Hahner $^1$\\
{\tt\scriptsize \href{mailto:mhahner@vision.ee.ethz.ch}{mhahner@vision.ee.ethz.ch}}
\and
Christos Sakaridis $^1$\\
{\tt\scriptsize \href{mailto:csakarid@vision.ee.ethz.ch}{csakarid@vision.ee.ethz.ch}}
\and
Dengxin Dai $^{1,2}$\\
{\tt\scriptsize \href{mailto:ddai@mpi-inf.mpg.de}{ddai@mpi-inf.mpg.de}}
\and
Luc Van Gool $^{1,3}$\\
{\tt\scriptsize \href{mailto:luc.vangool@kuleuven.be}{luc.vangool@kuleuven.be}}
\and
$^{1}$ \text{ ETH Zürich}
\and
$^{2}$ \text{ MPI for Informatics}
\and
$^{3}$ \text{ KU Leuven}
}

\maketitle

\begin{abstract}
   This work addresses the challenging task of LiDAR-based 3D object detection in foggy weather. 
   Collecting and annotating data in such a scenario is very time, labor and cost intensive. 
   In this paper, we tackle this problem by simulating physically accurate fog into clear-weather scenes, so that the abundant existing real datasets captured in clear weather can be repurposed for our task. 
   Our contributions are twofold:
   1) We develop a physically valid fog simulation method that is applicable to any LiDAR dataset. This unleashes the acquisition of large-scale foggy training data at no extra cost. These partially synthetic data can be used to improve the robustness of several perception methods, such as 3D object detection and tracking or simultaneous localization and mapping, on real foggy data.
   2) Through extensive experiments with several state-of-the-art detection approaches, we show that our fog simulation can be leveraged to significantly improve the performance for 3D object detection in the presence of fog. Thus, we are the first to provide strong 3D object detection baselines on the Seeing Through Fog dataset. Our code is available at \href{https://trace.ethz.ch/lidar_fog_simulation}{www.trace.ethz.ch/lidar\_fog\_simulation}.
\end{abstract}

\section{Introduction}
\label{sec:intro}

\begin{figure}
     \centering
     \begin{subfigure}[b]{\linewidth}
         \centering
         \includegraphics[width=\linewidth]{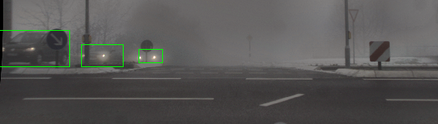}
     \end{subfigure}
     \begin{subfigure}[b]{0.495\linewidth}
         \centering
         \includegraphics[width=\linewidth]{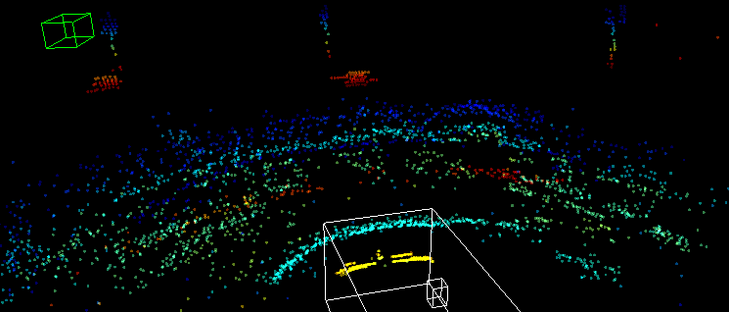}
         \caption{\textit{strongest} returns}
         \label{fig:fog_front_strongest}
     \end{subfigure}
     \hfill
     \begin{subfigure}[b]{0.495\linewidth}
         \centering
         \includegraphics[width=\linewidth]{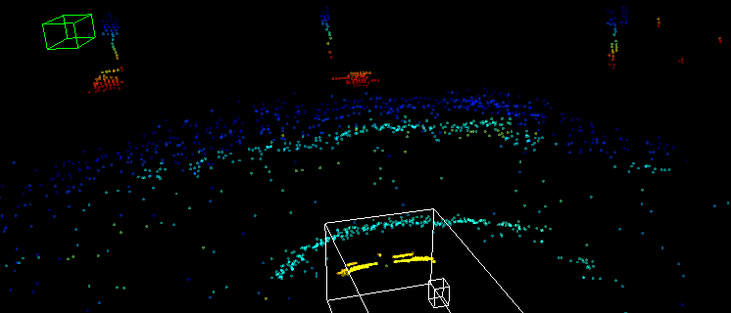}
         \caption{\textit{last} returns}
         \label{fig:fog_front_last}
     \end{subfigure}
     \caption{LiDAR returns caused by fog in the (top) scene. (a) shows the \textit{strongest} returns and (b) the \textit{last} returns, color coded by the LiDAR \textit{channel}. 
     The returns of the ground are removed for better visibility of the points introduced by fog.
     Best viewed in color (\textcolor{red}{red} $\widehat{=}$ low, \textcolor{cyan}{cyan} $\widehat{=}$ high, 
     3D bounding box annotation in \textcolor{green}{green}, ego vehicle dimensions in \textcolor{gray}{gray}).}
     \label{fig:fog_front}
\end{figure}

Light detection and ranging (LiDAR) is crucial for the implementation of \textit{safe} autonomous cars, because LiDAR measures the precise distance of objects from the sensor, which cameras cannot measure directly. Thus, LiDAR has found its way into many applications, including detection~\cite{PP, PV-RCNN}, tracking~\cite{tracking, robotics8030075}, localization~\cite{localization, survey}, and mapping~\cite{LOAM, SLAM}. Despite the benefit of measuring exact depth information, LiDAR has a significant drawback. The light pulses that LiDAR sensors emit in the invisible near infrared (NIR) spectrum (typically at $850$ and $903$ to $905$ nm wavelength~\cite{LIBRE}) do \textit{not} penetrate water particles, as opposed to automotive radars. This means as soon as there are water particles in the form of fog in the air, light pulses emitted by the sensor will undergo backscattering and attenuation. Attenuation reduces the received signal power that corresponds to the range of the solid object in the line of sight which should be measured, while backscattering creates a spurious peak in the received signal power at an incorrect range. As a result, the acquired LiDAR point cloud will contain some spurious returns whenever there is fog present at the time of capture. This poses a big challenge for most outdoor applications, as they typically require robust performance under all weather conditions.

In recent years, several LiDAR datasets for 3D object detection~\cite{KITTI, nuscenes2019, lyft2019, Argoverse, WaymoOD, Honda3D, A2D2, Apollo} have been presented. Although many of them contain diverse driving scenarios, none of them allows an evaluation on different kinds of adverse weather. Only recently, the Canadian Adverse Driving Conditions (CADC) Dataset~\cite{CADC} and the Seeing Through Fog (STF) Dataset~\cite{STF} address the need for such an evaluation. While CADC focuses on snowfall, STF is targeted towards \textit{evaluation} under fog, rain and snow. Consequently, there is still not a large quantity of LiDAR foggy data available that could be used to \textit{train} deep neural networks. 

The reason for this is obvious: collecting and annotating large-scale datasets per se is time, labor and cost intensive, let alone when done for adverse weather conditions.

This is exactly the shortfall that our work addresses. In Sec.~\ref{sec:derivation}, we propose a physically-based fog simulation that converts real clear-weather LiDAR point clouds into \textit{foggy} counterparts. In particular, we use the standard linear system~\cite{Rasshofer_2011} that models the transmission of LiDAR pulses. We distinguish between the cases of clear weather and fog with respect to the impulse response of this system and establish a formal connection between the received response under fog and the respective response in clear weather. This connection enables a straightforward transformation of the range and intensity of each original clear-weather point, so that the new range and intensity correspond to the measurement that would have been made if fog was present in the scene. We then show in Sec.~\ref{sec:results} that several state-of-the-art 3D object detection pipelines can be trained on our partially synthetic data to get improved robustness on \textit{real} foggy data. This scheme has already been applied on images for semantic segmentation~\cite{SDV18, SynRealDataFogECCV18, FoggySynscapes} and we show that it is also successful for LiDAR data and 3D object detection. 

For our experiments, we simulate fog on the clear-weather training set of STF~\cite{STF} and evaluate on their real foggy test set. Fig.~\ref{fig:fog_front} shows an example scene from the STF dense fog test set, where the noise introduced by fog is clearly visible in the LiDAR data. The authors of STF~\cite{STF} used a Velodyne HDL-64E as their main LiDAR sensor. This sensor comes with 64 \textit{channels} and a so-called \textit{dual mode}. In this mode, it can measure not only the \textit{strongest}, but also the \textit{last} return received for each individual emitted light pulse. 
Even though the \textit{last} signal contains less severe noise, fog still causes a significant amount of spurious returns. Therefore, even in this \textit{dual mode}, the sensor cannot fully ``see through fog''.

\begin{figure}
     \centering
     \begin{subfigure}[b]{\linewidth}
         \centering
         \includegraphics[width=\linewidth]{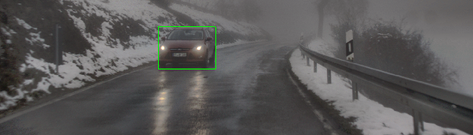}
     \end{subfigure}
     \begin{subfigure}[b]{0.495\linewidth}
         \centering
         \includegraphics[width=\linewidth]{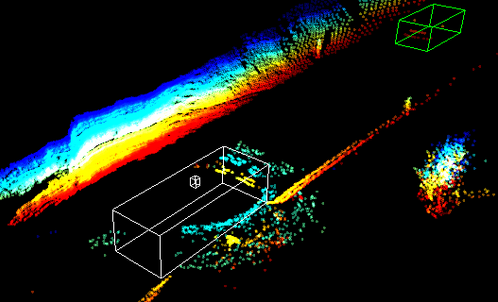}
         \caption{\textit{channel}}
         \label{fig:fog_halfcircle_channel}
     \end{subfigure}
     \hfill
     \begin{subfigure}[b]{0.49\linewidth}
         \centering
         \includegraphics[width=\linewidth]{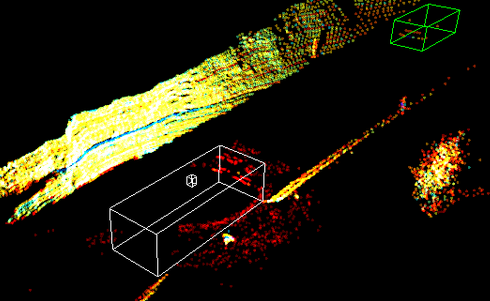}
         \caption{\textit{intensity}}
         \label{fig:fog_halfcircle_intensity}
     \end{subfigure}
     \caption{LiDAR returns caused by fog in the (top) scene. 
     Color coded by the LiDAR \textit{channel} in (a) and by the \textit{intensity} in (b). 
     The returns of the ground are removed for better visibility of the points introduced by fog.
     Best viewed in color, same color coding as in Fig.~\ref{fig:fog_front} applies.}
     \label{fig:fog_halfcircle}
\end{figure}

Fig.~\ref{fig:fog_halfcircle} shows an interesting characteristic of the noise introduced by fog, namely that it is not uniformly distributed around the sensor. On the contrary, the presence of noise depends on whether there is a target in the line of sight below a certain range from the sensor. If there is a solid object at a moderate range, there are few, if any, spurious returns from the respective pulses. On the other hand, if there is no target in the line of sight below a certain range, there are a lot of spurious returns that are caused by fog. This becomes apparent in the example of Fig.~\ref{fig:fog_halfcircle}, where on the left side of the road there is a hill and on the right side there is open space behind the guardrail. Only in the latter case does the noise caused by fog appear in the measurement. This behavior is explained with our theoretical formulation in Sec.~\ref{sec:derivation}.


As a side note, similar sensor noise can also be caused by exhaust smoke, but if the future of transportation goes electric, at least this problem may \textit{vanish into thin air}.

\section{Related Work}
\label{sec:related}

\subsection{Effects of Adverse Weather on LiDAR}
\label{sec:related_lidar}
Some of the early works include Isaac \etal~\cite{Isaac}. In 2001, they investigate the influences of fog and haze on optical wireless communications in the NIR spectrum. Then, in 2011, Rasshofer \etal~\cite{Rasshofer_2011} investigate the influences of weather phenomena on automotive LiDAR systems. In recent years, adverse weather conditions got a lot more attention and there are many other works worth mentioning that look into the degradation of LiDAR data in different adverse weather conditions~\cite{Wojtanowski, Filgueira, Kutila, Jokela, Heinzler, Wallace, Li, Kutila2}. Very recently, in 2020, the authors of LIBRE~\cite{LIBRE} test several LiDAR sensors in a weather chamber under rain and fog. Thereby they provide great and valuable insights on the robustness of individual sensors of this time on challenging weather conditions.

\subsection{Adverse Weather and LiDAR Simulation}
\label{sec:related_simulation}

In the automotive context, artificial fog simulation is so far mostly limited to image based methods. Sakaridis \etal~\cite{SDV18} e.g.\ create a foggy version of Cityscapes~\cite{Cityscapes}, a dataset for Semantic Segmentation, and Hahner \etal~\cite{FoggySynscapes} a foggy version of the purely synthetic dataset Synscapes~\cite{Synscapes} by leveraging the depth information given in the original datasets. Sakaridis \etal also released ACDC~\cite{ACDC}, a dataset providing semantic pixel-level annotations for 19 Cityscapes classes in adverse conditions. Only recently, Bijelic \etal~\cite{STF} propose a first order approximation to simulate fog in an automotive LiDAR setting. However, their simulation only aims at reproducing measurements they carried out in a 30m long fog chamber. 

Goodin \etal~\cite{Goodin} develop a model to quantify the performance degradation of LIDAR in rain and incorporate their model into a simulation which they use to test an advanced driver assist system (ADAS). 
Michaud \etal~\cite{Michaud2015TowardsCT} and Tremblay \etal~\cite{Tremblay} propose a method to render rain on images to evaluate and improve the robustness on rainy images. 

\subsection{3D Object Detection}
\label{sec:related_detection}

After the release of many LiDAR datasets~\cite{KITTI, nuscenes2019, lyft2019, Argoverse, WaymoOD, Honda3D, A2D2, Apollo, CADC, STF} over the past few years, 3D object detection is receiving increasing attention in the race towards autonomous driving. While there exist camera based methods such as Simonelli \etal ~\cite{Simonelli_2019_ICCV} and gated camera methods such as Gated3D~\cite{Gated3D}, the top ranks across all dataset leaderboards are typically sorted out among LiDAR based methods.

Seminal work in LiDAR based 3D object detection methods include PointNet~\cite{PointNet} and VoxelNet~\cite{VoxelNet}. PointNet~\cite{PointNet} is a neural network that directly processes point clouds without quantizing the 3D space into voxels beforehand. Most notably, the network architecture is robust to input perturbation, so the order in which the points get fed into the network do not effect its performance. VoxelNet~\cite{VoxelNet} builds on the idea to quantize the 3D space into equally sized voxels and then leverages PointNet-like layers to process each and every voxel. Due to it's computational intensive 3D convolutions, it is however a rather heavy architecture. 

That's where PointPillars~\cite{PP} comes in: it gets rid of the quantization in the height domain and processes the point cloud instead in a 3D pillar grid. PointPillars~\cite{PP} is based on the SECOND~\cite{SECOND} codebase, but due to the novel pillar idea, it can fall back to much faster 2D convolutions and achieves very competitive results at a much faster speed. Its successor PointPainting~\cite{PointPainting} further leverages image segmentation results and ``paints'' the points with a pseudo class label before processing them with the PointPillars~\cite{PP} architecture. 

Shi \etal achieved several recent milestones in 3D object detection. PointRCNN~\cite{PRCNN} is a two-stage architecture, where the first stage generates 3D bounding box proposals from a point cloud in a bottom-up manner and the second stage refines these 3D bounding box proposals in a canonical fashion. Part-A$^2$~\cite{PartA2} is part-aware in a sense that the network takes into account which part of the object a point belongs to. It leverages these intra-object part locations and can thereby achieve higher results. PV-RCNN~\cite{PV-RCNN} and it's successor PV-RCNN++~\cite{shi2021pvrcnn} are the latest of their works that simultaneously process (coarse) voxels and the raw points of the point cloud at the same time.
\section{Fog Simulation on Real LiDAR Point Clouds}
\label{sec:derivation}

To simulate the effect of fog on real-world LiDAR point clouds that have been recorded in clear weather, we need to resort to the optical system model that underlies the function of the transmitter and receiver of the LiDAR sensor. In particular, we examine a single measurement/point, model the full signal of received power as a function of the range and recover its exact form corresponding to the original clear-weather measurement. This allows us to operate in the signal domain and implement the transformation from clear weather to fog simply by modifying the part of the impulse response that pertains to the optical channel (i.e.\ the atmosphere). In the remainder of this section, we first provide the background on the LiDAR sensor's optical system and then present our fog simulation algorithm based on this system. 

\subsection{Background on the LiDAR Optical Model}
\label{sec:derivation:background}

Rasshofer \etal~\cite{Rasshofer_2011} introduced a simple linear system model to describe the received signal power at a LiDAR's receiver, which is valid for non-elastic scattering. In particular, the range-dependent received signal power $P_R$ is modeled as the time-wise convolution between the time-dependent transmitted signal power $P_T$ and the time-dependent impulse response $H$ of the environment:
\begin{equation} \label{eq:LiDAR:equation}
    P_R(R) = C_A \int_0^{2R/c} P_T(t)\,H\left(R-\frac{ct}{2}\right) dt.
\end{equation}
$c$ is the speed of light and $C_A$ is a system constant which is independent of time and range. For our fog simulation, as we explain in Sec.~\ref{sec:derivation:simulation}, $C_A$ can be factored out.

We proceed with the description and modeling of the remaining terms in \eqref{eq:LiDAR:equation}. The time signature of the transmit pulse can be modeled for automotive LiDAR sensors~\cite{Rasshofer_2011} by a $\sin^2$ function:
\begin{equation} \label{eq:pulse:sin2}
    P_T(t) = \left\{
    \begin{array}{rl}
        P_0\sin^2\left(\frac{\pi}{2\,\tau_H}t\right), & 0 \leq t \leq 2\,\tau_H \\
        0 & \text{otherwise}.
    \end{array}
    \right.
\end{equation}
Where $P_0$ denotes the pulse's peak power and $\tau_H$ the half-power pulse width. Typical values for $\tau_H$ lie between 10 and 20 ns~\cite{Rasshofer_2011}. In \eqref{eq:pulse:sin2}, the time origin is set to the start of the pulse, so in case a LiDAR sensor does not report the range associated with a rising edge, but the maximum of the corresponding peak in the return signal, we can perform the required correction later in our pipeline. Since it is common to report the rising edge in embedded signal processing, we keep this convention throughout all of our equations and show where one can perform such a correction later on. 

\begin{figure}
     \centering
     \includegraphics[width=\linewidth]{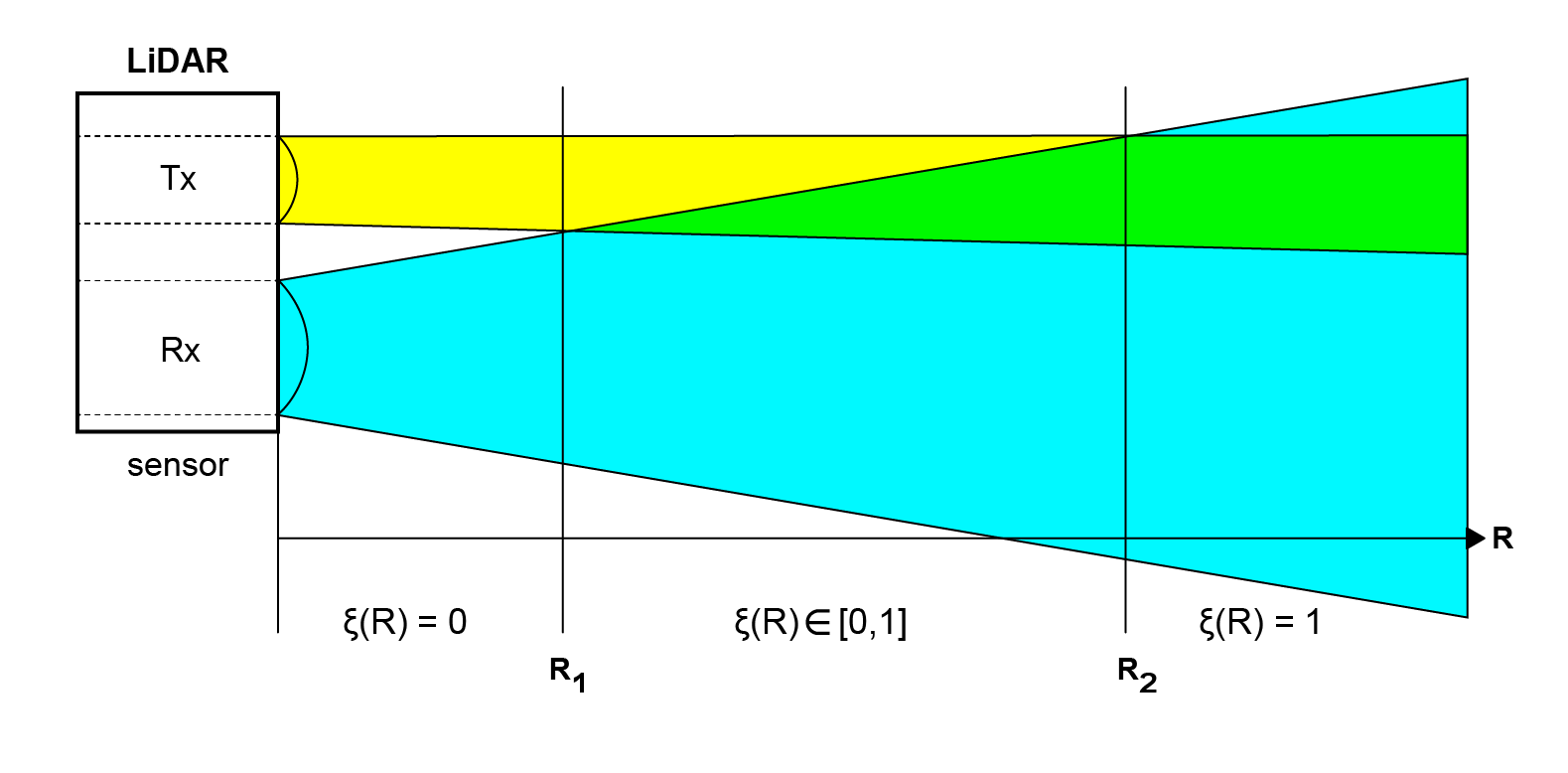}
     \caption{Sketch of a LiDAR sensor where the transmitter \textsf{Tx} and the receiver \textsf{Rx} do not have coaxial optics, but have parallel axes. This is called a bistatic beam configuration. Figure adjusted from~\cite{Rasshofer_2011}.}
     \label{fig:xi}
\end{figure}

The spatial impulse response function $H$ of the environment can be modeled as the product of the individual impulse responses of the optical channel, $H_C$, and the target, $H_T$:
\begin{equation} \label{eq:impulse:response:decomposition}
    H(R) = H_C(R)\,H_T(R).
\end{equation}
The impulse response of the optical channel $H_C$ is
\begin{equation} \label{eq:impulse:response:channel}
    H_C(R) = \frac{T^2(R)}{R^2} \xi(R),
\end{equation}
where $T(R)$ stands for the total one-way transmission loss and $\xi(R)$ denotes the crossover function defining the ratio of the area illuminated by the transmitter and the part of it observed by the receiver, as illustrated in Fig.~\ref{fig:xi}. Because generally the full details of the optical configuration in commercial LiDAR sensors are not publicly available (i.e.\ the precise values of $R_1$ and $R_2$ are unknown), $\xi(R)$ in our case is a piece-wise linear approximation defined as
\begin{equation} \label{eq:xi}
    \xi(R) = \left\{
    \begin{array}{rl}
        0, & R \leq R_1 \\
        \frac{R}{R_2-R_1} -\frac{R_1}{R_2-R_1}, & R_1 < R < R_2 \\
        1, & R_2 \leq R.
    \end{array}
    \right.
\end{equation}
The total one-way transmission loss $T(R)$ is defined as 
\begin{equation} \label{eq:transmission:loss}
    T(R) = \exp \left(-\int_{r=0}^{R} \alpha(r)dr \right),
\end{equation}
where $\alpha(r)$ denotes the spatially varying attenuation coefficient. In our simulation, we make the assumption of a \emph{homogeneous} optical medium, i.e.\ $\alpha(r) = \alpha$. As a result, \eqref{eq:transmission:loss} yields 
\begin{equation} \label{eq:transmission:loss:homogeneous}
    T(R) = \exp\left(-\alpha{}R\right).
\end{equation}
The attenuation coefficient $\alpha$ depends on the weather at the time of measurement and increases as visibility range decreases. Therefore, for the same 3D scene, the impulse response of the optical channel $H_C$ varies with visibility.

The last term of the optical system \eqref{eq:LiDAR:equation} that remains to be modeled is the impulse response of the target, $H_T$. However, we need to distinguish cases for $H_T$ according to the weather condition, as the composition of the target for the same 3D scene is different in fog than in clear weather. We make the contribution of constructing a direct relation between the response $P_R$ in clear weather and in fog for the same 3D scene and this relation enables us to simulate fog on real clear-weather LiDAR measurements.

\subsection{Fog Simulation for LiDAR}
\label{sec:derivation:simulation}

We now particularize the optical model of Sec.~\ref{sec:derivation:background} for the individual cases of clear weather and fog in terms of the impulse response terms, $H_C$ and $H_T$.

In clear weather, the attenuation coefficient $\alpha$ is 0, so
\begin{equation} \label{eq:impulse:responce:channel:clear}
    H_C(R) = \frac{\xi(R)}{R^2}.
\end{equation}
Moreover, the target in clear weather is comprised only of the solid object on which the LiDAR pulse is reflected. This type of target is called a \emph{hard} target~\cite{Rasshofer_2011}. 

The impulse response $H_T$ of a hard target at range $R_0$ is a Dirac delta function of the form
\begin{equation} \label{eq:impulse:response:target:clear}
    H_T(R) = H_T^{\text{hard}}(R) = \beta_0 \delta(R-R_0),
\end{equation}
where $\beta_0$ denotes the differential reflectivity of the target. If we only consider diffuse reflections (Lambertian surfaces), $\beta_0$ is given by 
\begin{equation} \label{eq:beta_0}
    \beta_0 = \frac{\Gamma}{\pi},\;0 < \Gamma \leq 1.
\end{equation}

Consequently, plugging in \eqref{eq:impulse:responce:channel:clear} and \eqref{eq:impulse:response:target:clear} into \eqref{eq:impulse:response:decomposition}, in clear weather the total impulse response function $H(R)$ can be expressed as
\begin{equation} \label{eq:impulse:response:total:clear}
    H(R) = \frac{\xi(R_0)}{{R_0}^2} \beta_0 \delta(R-R_0),
\end{equation}
where we have used the property $f(x)\delta(x-x_0)=f(x_0)\delta(x-x_0)$. Since in practice $R_2$ is less than two meters~\cite{Velodyne}, we can safely assume $R_2 \ll R_0$, so $\xi(R_0) = 1$. Thus, starting from \eqref{eq:LiDAR:equation} and given the range measurement $R_0$ for the original clear-weather LiDAR point, we compute the received signal power in closed form as

{\scriptsize
\begin{align}
    &P_{R,\text{clear}}(R) \nonumber \\
    &{=}\;C_A \displaystyle\int_0^{2\tau_H} P_0 \sin^2\left(\frac{\pi}{2\tau_H}t\right) \frac{1}{{R_0}^2} \beta_0 \delta(R-\frac{ct}{2}-R_0) dt \nonumber \\
    &{=}\;\left\{\begin{array}{rl}
        \frac{C_A P_0 \beta_0}{{R_0}^2} \sin^2\left(\frac{\pi(R-R_0)}{c\tau_H}\right), & R_0 \leq R \leq R_0+c\tau_H \\
        0 & \text{otherwise}.
    \end{array}\right. \label{eq:response:clear}
\end{align}
}%
The received signal power attains its maximum value at $R_0+\frac{c\tau_H}{2}$. So as we mentioned in Sec.~\ref{sec:derivation:background}, here one can simply shift the response $P_{R,\text{clear}}(R)$ by $-\frac{c\tau_H}{2}$ if necessary.

We now establish the transformation of $P_{R,\text{clear}}(R)$ to $P_{R,\text{fog}}(R)$ under fog. While the same hard target still exists since the 3D scene is the same, there is now an additional contribution from fog---which constitutes a \emph{soft} target~\cite{Rasshofer_2011} that provides distributed scattering---to the impulse response $H_T$. 

The impulse response of this soft fog target, $H_T^{\text{soft}}$, is a Heaviside function of the form
\begin{equation} \label{eq:impulse:response:target:fog:soft}
    H_{T}^{\text{soft}}(R) = \beta{}U(R_0-R),
\end{equation}
where $\beta$ denotes the backscattering coefficient, which is constant under our homogeneity assumption, and $U$ is the Heaviside function. 

The co-existence of a hard target and a soft target can be modeled by taking the \emph{superposition} of the respective impulse responses:
\begin{align}
    H_T(R)\;&{=}\;H_T^{\text{soft}}(R) + H_T^{\text{hard}}(R) \nonumber\\
    &{=}\;\beta{}U(R_0-R) + \beta_0\delta(R-R_0). \label{eq:impulse:response:target:fog}
\end{align}

Consequently, plugging in \eqref{eq:transmission:loss:homogeneous} into \eqref{eq:impulse:response:channel} and then \eqref{eq:impulse:response:channel} and  \eqref{eq:impulse:response:target:fog} into \eqref{eq:impulse:response:decomposition}, in fog the total impulse response function $H(R)$ can be expressed as
\begin{align} \label{eq:impulse:response:total:fog}
    H(R) = &{\frac{\exp(-2\alpha{}R)\xi(R)}{R^2}} \nonumber \\ 
    &{\times}\;\left(\beta{}U(R_0-R) + \beta_0\delta(R-R_0)\right).
\end{align}

We observe that compared to clear weather, the spatial impulse response in fog is more involved, but it can still be decomposed into two terms, corresponding to the hard and the soft target respectively, leading to a respective decomposition of the received response as
\begin{equation} \label{eq:response:fog:decomposition}
    P_{R,\text{fog}}(R) = P_{R,\text{fog}}^{\text{hard}}(R) + P_{R,\text{fog}}^{\text{soft}}(R).
\end{equation}
Focusing on the hard target term, using \eqref{eq:LiDAR:equation} to calculate the corresponding term of the received response $P_{R,\text{fog}}^{\text{hard}}$ and leveraging again the assumption that $R_2 \ll R_0$, we obtain 
{\scriptsize
\begin{align}
    &{P^{\text{hard}}_{R,\text{fog}}}(R) = \nonumber \\
    &{=}\;C_A \frac{\exp(-2\alpha{}R_0)}{{R_0}^2} \displaystyle\int_0^{2\tau_H} P_0 \sin^2\left(\frac{\pi}{2\tau_H}t\right)\beta_0\delta(R-\frac{ct}{2}-R_0) dt \nonumber \\
    &{=}\;\left\{\begin{array}{rl}
        \frac{C_A P_0 \beta_0 \exp(-2\alpha{}R_0)}{{R_0}^2} \sin^2\left(\frac{\pi(R-R_0)}{c\tau_H}\right), & R_0 \leq R \leq R_0+c\tau_H \\
        0 & \text{otherwise}.
    \end{array}\right. \nonumber \\
    &{=}\;\exp(-2\alpha{}R_0)P_{R,\text{clear}}(R). \label{eq:response:fog:hard}
\end{align}
}%
In other words, the hard target term of the response in fog is an attenuated version of the original clear-weather response $P_{R,\text{clear}}$. On the other hand, the soft target term is 
{\scriptsize
\begin{align}
    {P^{\text{soft}}_{R,\text{fog}}}(R)\;&{=}\;{C_A} P_0 \beta \displaystyle\int_0^{2\tau_H} \sin^2\left(\frac{\pi}{2\tau_H}t\right) \times \nonumber \\ &{\times}\:\frac{\exp\Big(-2\alpha{}\left(R-\frac{ct}{2}\right)\Big)}{{\left(R-\frac{ct}{2}\right)}^2} \xi(R-\frac{ct}{2}) U(R_0-R+\frac{ct}{2}) dt \label{eq:response:fog:soft}
\end{align}
}%
and does not possess a closed-form expression, as the respective integral $I(R,R_0,\alpha,\tau_H)$ on the right-hand side of \eqref{eq:response:fog:soft} cannot be calculated analytically. 

\begin{figure}
     \centering
     \begin{subfigure}[b]{0.49\linewidth}
         \centering
         \includegraphics[width=\linewidth]{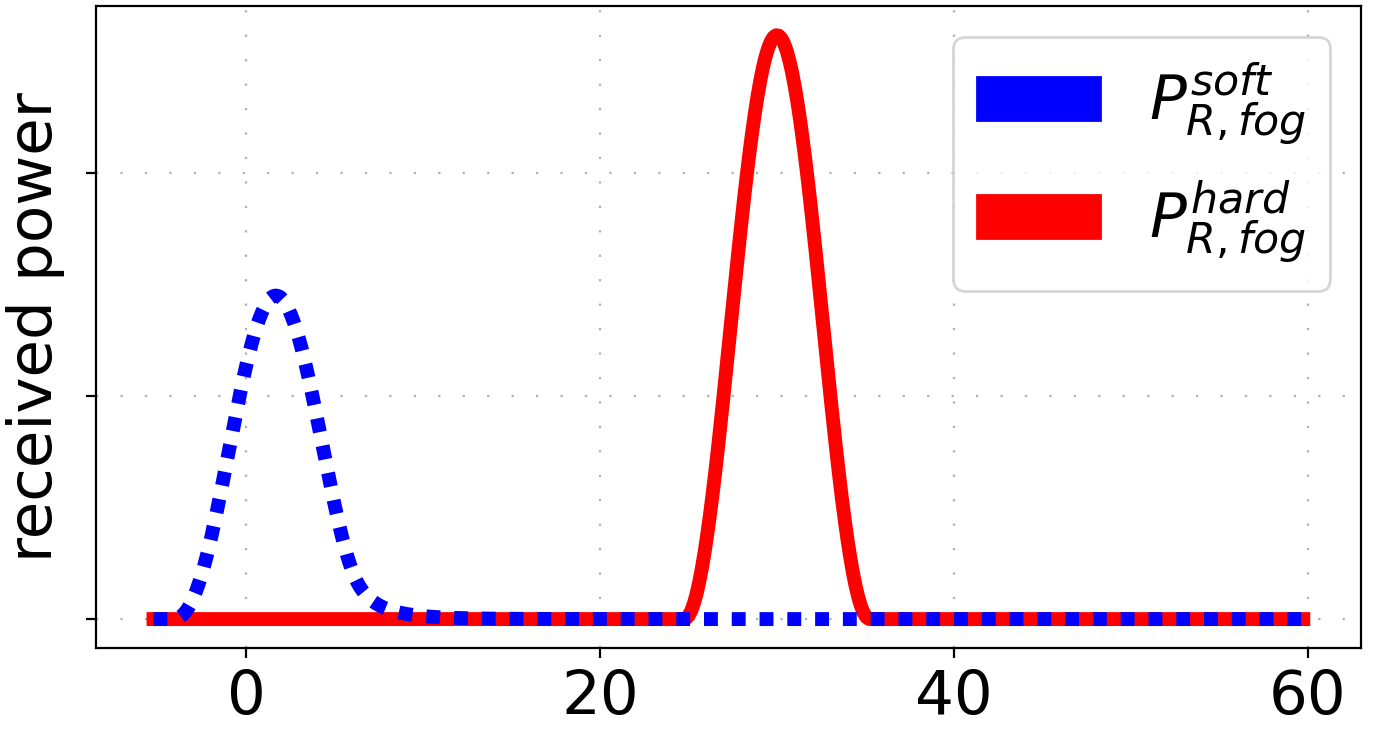}
         \caption{$\color{blue} P^{\text{soft}}_{R,\text{fog}}(R) 
         \color{black} < \color{red} P^{\text{hard}}_{R,\text{fog}}$}
         \label{fig:theory_hard}
     \end{subfigure}
     \hfill
     \begin{subfigure}[b]{0.49\linewidth}
         \centering
         \includegraphics[width=\linewidth]{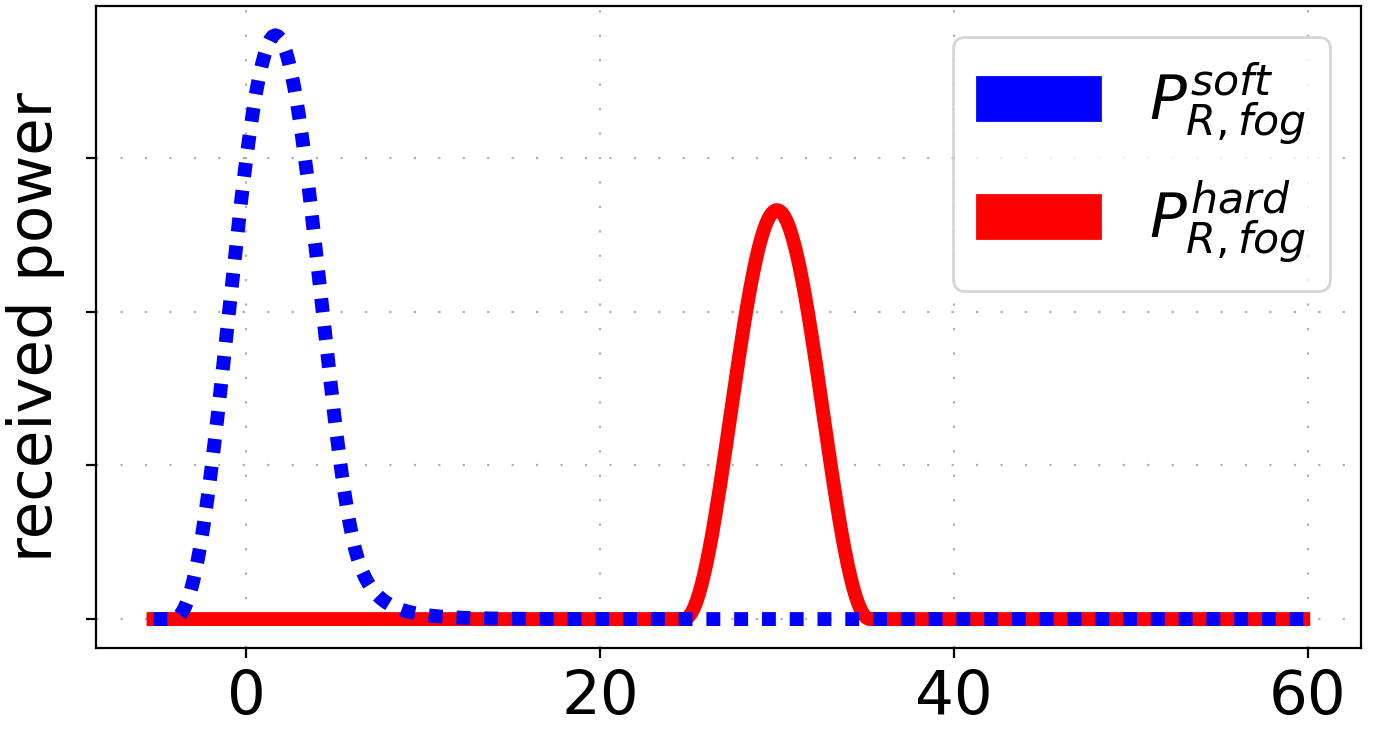}
         \caption{$\color{blue} P^{\text{soft}}_{R,\text{fog}} 
         \color{black} > \color{red} P^{\text{hard}}_{R,\text{fog}}$}
         \label{fig:theory_soft}
     \end{subfigure}
     \caption{The two terms of the received signal power $P_{R,\text{fog}}$ from a single LiDAR pulse, associated to the solid object that reflects the pulse ($P_{R,\text{fog}}^{\text{hard}}$) and the soft fog target ($P_{R,\text{fog}}^{\text{soft}}$), plotted across the range domain. While in (a) the fog is not thick enough to yield a return, in (b) it is thick enough to yield a return that overshadows the solid object at $R_0$\,=\,$30\text{m}$.}
     \label{fig:theory}
\end{figure}

However, for given $\tau_H$ and $\alpha$, $I(R,R_0,\alpha,\tau_H)$ can be computed numerically for fixed values of $R$. We use Simpson's \sfrac{1}{3} rule for numerical integration and provide indicative examples of the profile of $P^{\text{soft}}_{R,\text{fog}}(R)$ in Fig.~\ref{fig:theory}. Depending on the distance of the hard target from the sensor, the soft target term of the response may exhibit a larger maximum value than the hard target term, which implies that \emph{the measured range changes due to the presence of fog} and becomes equal to the point of maximum of the soft-target term.

The formulation that we have developed affords a simple algorithm for fog simulation on clear-weather point clouds. The input parameters to the algorithm are $\alpha$, $\beta$, $\beta_0$ and $\tau_H$. The main input of the algorithm is a clear-weather point cloud, where each point $\mathbf{p} \in \mathbb{R}^3$ has a measured intensity $i$. We make the assumption that the intensity readings of the sensor are a linear function of the maxima of the received signal power $P_{R,\text{clear}}$ corresponding to each measurement. The procedure for each point $\mathbf{p}$ is given in Algorithm~\ref{algorithm}. Note, that we add some noise to the distance of $P_{R,\text{fog}}^{\text{soft}}$ (line 14-15), otherwise all points introduced by $P_{R,\text{fog}}^{\text{soft}}$ would lie precisely on a circle around the LiDAR sensor.


\begin{algorithm}
  \caption{LiDAR fog simulation}\label{algorithm}
  \begin{algorithmic}[1]
    \Procedure{foggify}{$\mathbf{p}$, $i$, $\alpha$, $\beta$, $\beta_0$, $\tau_H$}
      \State $R_0 \gets \|\mathbf{p}\|$
      \State $x, y, z \gets \mathbf{p}$                                              \scriptsize \Comment{$i = P_{R,\text{clear}}$} \normalsize
      \State $C_A P_0 \gets i\frac{{R_0}^2}{\beta_0}$                                   \scriptsize \Comment{follows from Eq.~\eqref{eq:response:clear}} \normalsize
      
      \State $i_{\text{hard}} \gets i \times \exp(-2\alpha{}R_0)$                       \scriptsize \Comment{see Eq.~\eqref{eq:response:fog:hard}} \normalsize
      
      \For{$R$ in $(0, 0.1, ..., R_0)$}                                                 \scriptsize \Comment{\scriptsize $10$cm accuracy} \normalsize
        \State $I_R \gets$ \scriptsize SIMPSON\normalsize($I(R, R_0,\alpha,\tau_H))$    \scriptsize \Comment{see Eq.~\eqref{eq:response:fog:soft}} \normalsize
      \EndFor
      
      \State $i_{\text{tmp}} \gets \max(I_R)$
      \State $R_{\text{tmp}} \gets \argmax(I_R)$
      
      \State $i_{\text{soft}} \gets C_A P_0 \,\beta \times i_{\text{tmp}}$              \scriptsize \Comment{see again Eq.~\eqref{eq:response:fog:soft}} \normalsize
      
      \If{$i_{\text{soft}} > i_{\text{hard}} $}
        \State $s \gets \frac{R_{\text{tmp}}}{R_0}$                                     \scriptsize \Comment{scaling factor $s$} \normalsize
        \State $p \gets$ \scriptsize RANDOM\_UNIFORM\_FLOAT\normalsize$(-1, 1)$                    
        \State $n \gets 2^p$                                                            \scriptsize \Comment{noise factor $n \in (\frac{1}{2}, 2)$} \normalsize
        \State $x \gets s \times n \times x$
        \State $y \gets s \times n \times y$
        \State $z \gets s \times n \times z$
        \State $i \gets i_{\text{soft}}$
      \Else                                                                             \scriptsize \Comment{keep original location} \normalsize
        \State $i \gets i_{\text{hard}}$                                                \scriptsize \Comment{only modify intensity} \normalsize
      \EndIf
      
      \State \textbf{return} $x, y, z, i$                                               
    \EndProcedure
  \end{algorithmic}
\end{algorithm}

\begin{figure}
     \centering
     \begin{subfigure}[b]{0.495\linewidth}
         \centering
         \includegraphics[width=\linewidth]{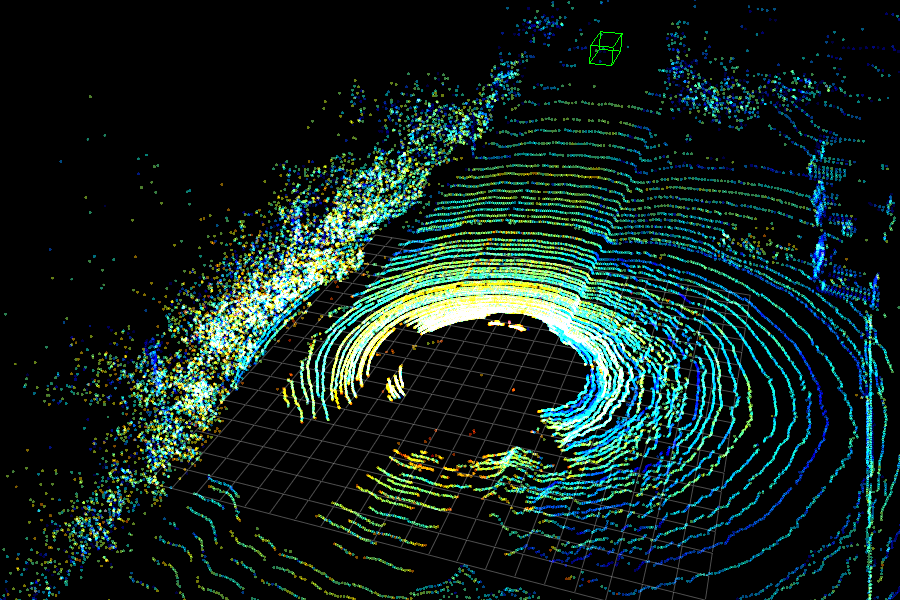}
     \end{subfigure}
     \hfill
     \begin{subfigure}[b]{0.495\linewidth}
         \centering
         \includegraphics[width=\linewidth]{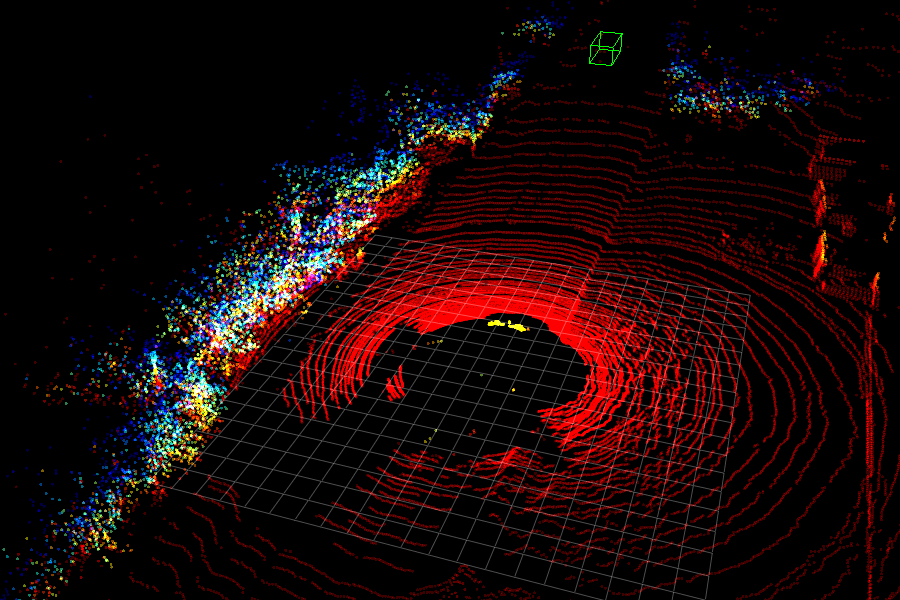}
     \end{subfigure}
     \begin{subfigure}[b]{0.495\linewidth}
         \centering
         \includegraphics[width=\linewidth]{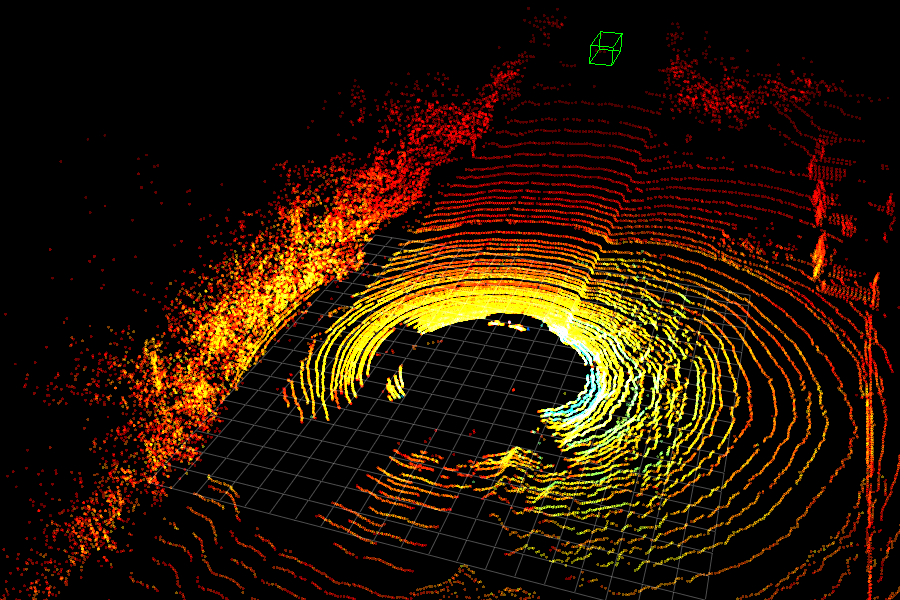}
     \end{subfigure}
     \hfill
     \begin{subfigure}[b]{0.495\linewidth}
         \centering
         \includegraphics[width=\linewidth]{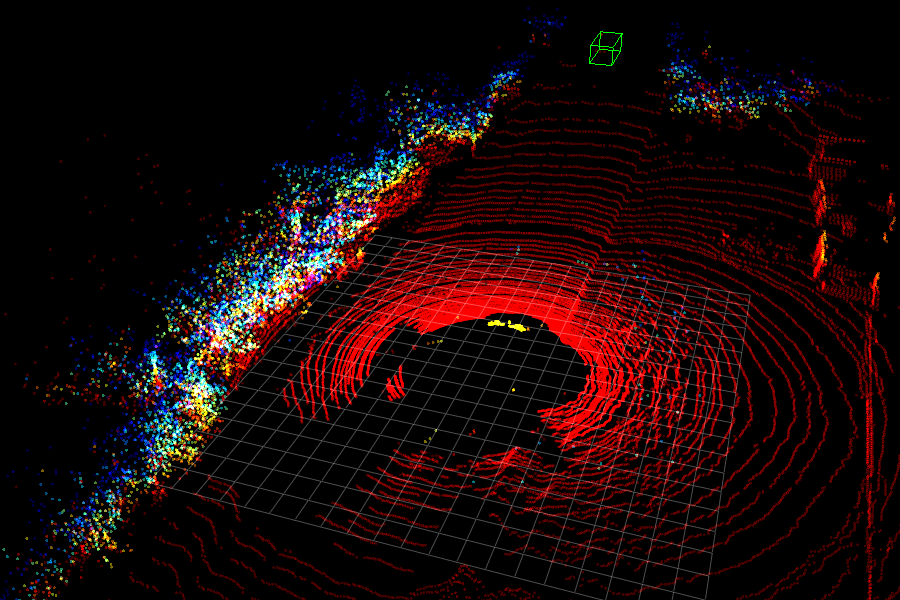}
     \end{subfigure}
     \begin{subfigure}[b]{0.495\linewidth}
         \centering
         \includegraphics[width=\linewidth]{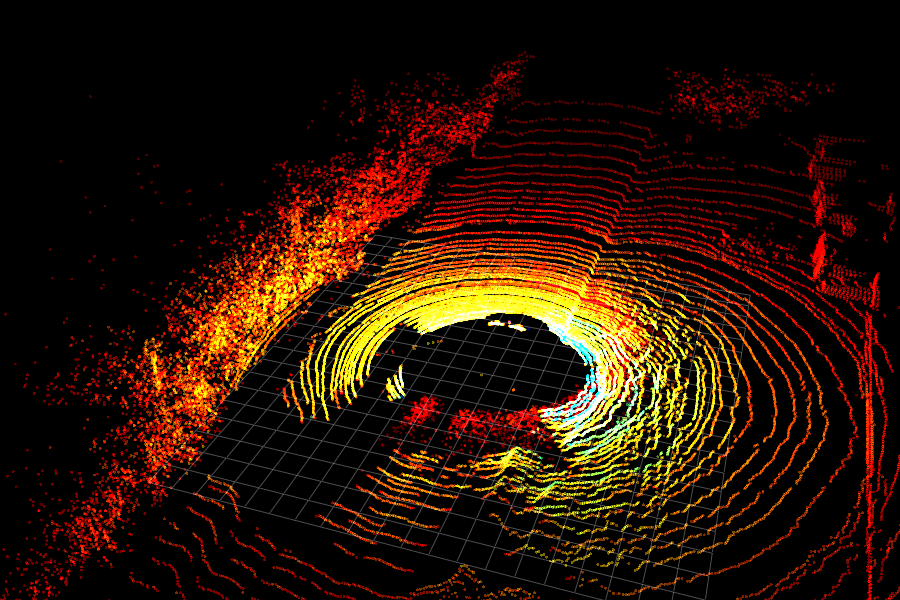}
     \end{subfigure}
     \hfill
     \begin{subfigure}[b]{0.495\linewidth}
         \centering
         \includegraphics[width=\linewidth]{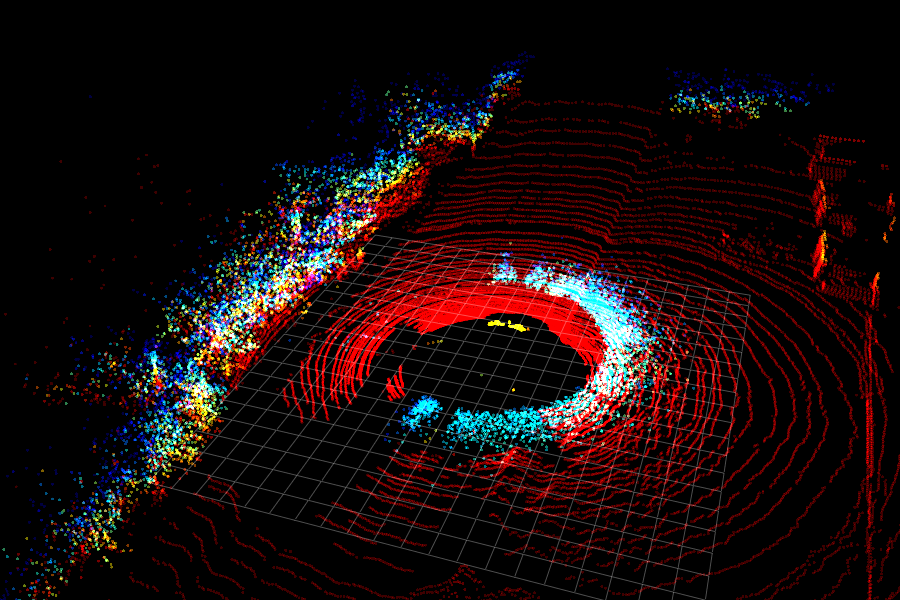}
     \end{subfigure}
     \caption{Comparison of our fog simulation (bottom) to the fog simulation in~\cite{STF} (middle) with $\alpha$ set to $0.06$, which corresponds to a meteorological optical range (MOR) $\approx50$m. 
     In the left column, the point cloud is color coded by the \textit{intensity} and in the right column it is color coded by the \textit{height} ($z$ value). 
     The top row shows the original point cloud.}
     \label{fig:fog_comparison}
\end{figure}

\section{Results}
\label{sec:results}

\subsection{Fog Simulation}
\label{sec:fog_results}

A qualitative comparison between our fog simulation and the fog simulation in~\cite{STF} can be found in Fig.~\ref{fig:fog_comparison}.  We can see that in contrast to the fog simulation in~\cite{STF} where the response of soft target is only modeled heuristically, our fog simulation models $P^{\text{soft}}_{R,\text{fog}}$ in a physically sound way. To highlight this difference, we specifically picked a clear weather scene with a similar layout to the real foggy scene depicted in Fig.~\ref{fig:fog_halfcircle}. Only in our fog simulation (best visible in the bottom right visualization of Fig.~\ref{fig:fog_comparison}), a similar half circle of fog noise gets simulated. In the Supplementary Materials we show a comparison with additional $\alpha$ values.

\subsection{3D Object Detection in Fog}
\label{sec:setup}

Our experimental setup codebase is forked from \hyperlink{https://github.com/open-mmlab/OpenPCDet}{OpenPCDet}~\cite{openpcdet2020}. 
It comes with implementations of the 3D Object Detection methods PV-RCNN~\cite{PV-RCNN}, PointRCNN~\cite{PRCNN}, SECOND~\cite{SECOND}, Part-A$^2$~\cite{PartA2} and PointPillars~\cite{PP}.
For our experiments we train all of these methods from scratch for $80$ epochs with their provided standard training policies on the STF~\cite{STF} dataset. 
We also tried to fine-tune from KITTI~\cite{KITTI} weights (which uses the same LiDAR sensor), but besides the networks converging faster, we did not see any benefit, so all the numbers you see in section~\ref{sec:setup} are trained from scratch on the STF~\cite{STF} clear weather training set that consist of $3469$ scenes.
The STF~\cite{STF} clear weather validation and testing set consists of $781$ and $1847$ scenes respectively. 
However, the main benefit of using STF~\cite{STF} for our experiments is because it comes with test sets for different adverse weather conditions. 
In particular, it comes with a light fog test set of $946$ scenes and a dense fog test set with $786$ scenes. 
This allows us to test the effectiveness of our fog simulation pipeline on real foggy data. 

Regarding our fog simulation, we assumed the half-power pulse width $\tau_H$ of the Velodyne HDL-64E sensor to be $20$ns and set $\beta$ to $\frac{0.046}{\text{MOR}}$ as in Rasshofer \etal~\cite{Rasshofer_2011}. We empirically set $\beta_0$ to $\frac{1 \times 10^{-6}}{\pi}$ for all points to get a similar intensity distribution as we can observe in the real foggy point clouds of STF~\cite{STF}. Since the Velodyne HDL-64E uses some unknown internal dynamic gain mechanism, it delivers at each and every time step intensity values in the full value range $[0, 255]$. To mimic this behaviour and also cover the full value range again we linearly scale up the intensity values after they have been modified by Algorithm~\ref{algorithm}. 

\begin{table*}
\customsize
\begin{tabular}{ cl rrr | rrr | rrr | rrr }
\multirow{2}{*}{Method} & \multirow{2}{*}{$\alpha$} & \multicolumn{3}{c|}{Car AP@.5IoU} & \multicolumn{3}{c|}{Cyclist AP@.25IoU} & \multicolumn{3}{c|}{Pedestrian AP@.25IoU}  & \multicolumn{3}{c}{mAP over classes}                                         \\ 

                                    &		& easy		    & mod.		    & hard		    & easy		    & mod.	    	& hard		    & easy		    & mod.		    & hard          & easy		    & mod.		    & hard              \\ 

\hline \noalign{\vskip 1mm} 

PV-RCNN~\cite{PV-RCNN} $\dagger$    & 0	    & 45.03         & 46.00         & 45.08         & 24.33         & 24.63         & 24.63         & 43.96         & 41.92         & 40.09         & 37.77         & 37.51         & 36.60             \\ 
PV-RCNN~\cite{PV-RCNN} $\ddagger$   & 0	    & 45.24         & 46.18         & 45.25         & 24.38         & 24.67         & 24.67         &\textbf{44.81} &\textbf{43.09} &\textbf{40.98} & 38.15         & 37.98         & 36.97             \\ 

\noalign{\vskip 1mm} 

fog simulation in \cite{STF}        & *     & 45.60	        & 46.60	        & 45.60	        & 26.42	        & 26.93	        & 27.80	        & 42.95	        & 40.89	        & 39.09         & 38.32         & 38.14         & 37.50             \\ 
our fog simulation                  & *     &\textbf{46.69}	&\textbf{47.38}	&\textbf{46.51}	&\textbf{27.89}	&\textbf{27.89}	&\textbf{29.29}	& 42.38	        & 40.65	        & 39.20         &\textbf{38.99}	&\textbf{38.64} &\textbf{38.33}     \\ 

\noalign{\vskip 1mm} \hline \noalign{\vskip 1mm} 

PointRCNN~\cite{PRCNN} $\dagger$    & 0	    & 44.00	        & 45.03	        & 43.73	        & 22.99         & 22.99         & 24.23	        & 41.73         & 38.38	        & 35.71	        & 36.24         & 35.47         & 34.56             \\ 
PointRCNN~\cite{PRCNN} $\ddagger$   & 0	    & 44.40         & 45.25         & 44.15         &\textbf{23.52} &\textbf{23.52} &\textbf{25.62} & 43.23         & 40.16         & 37.05         & 37.05         & 36.31         & 35.61             \\ 

\noalign{\vskip 1mm} 

fog simulation in \cite{STF}        & *     & 46.08         & 47.02         & 45.85         & 20.36         & 20.36         & 20.36         & 43.03         & 41.60         & 39.95         & 36.49         & 36.33         & 35.39             \\ 
our fog simulation                  & *     &\textbf{47.81} &\textbf{47.99} &\textbf{46.68} & 22.88         & 22.88         & 25.18         &\textbf{45.79} &\textbf{43.47} &\textbf{41.33} &\textbf{38.83} &\textbf{38.11} &\textbf{37.73}     \\ 

\noalign{\vskip 1mm} \hline \noalign{\vskip 1mm} 

SECOND~\cite{SECOND} $\dagger$      & 0	    & 42.36	        & 42.99	        & 41.99	        & 24.03         & 25.21         & 25.21	        & 36.72	        & 35.37	        & 33.84	        & 34.37         & 34.52         & 33.68             \\ 
SECOND~\cite{SECOND} $\ddagger$     & 0	    & 42.78         & 43.47         & 42.42         & 22.32         & 23.69         & 23.69         & 37.06         & 36.14         & 34.14         & 34.05         & 34.43         & 33.42             \\ 

\noalign{\vskip 1mm} 

fog simulation in \cite{STF}        & *     & 42.67         & 43.58         & 42.77         & 26.28         & 27.11         & 27.11         & 37.89         & 36.54         & 35.38         & 35.61         & 35.74         & 35.09             \\ 
our fog simulation                  & *     &\textbf{43.47} &\textbf{44.01} &\textbf{43.20} &\textbf{26.85} &\textbf{27.21} &\textbf{27.46} &\textbf{38.41} &\textbf{37.06} &\textbf{35.87} &\textbf{36.24} &\textbf{36.09} &\textbf{35.51}     \\ 

\noalign{\vskip 1mm} \hline \noalign{\vskip 1mm} 

Part-A²~\cite{PartA2} $\dagger$     & 0	    & 37.60	        & 38.15	        & 37.76	        & 24.51	        & 25.59	        & 25.59	        &\textbf{41.03}	&\textbf{39.29} &\textbf{37.59} & 34.38         & 34.34         & 33.65             \\ 
Part-A²~\cite{PartA2} $\ddagger$    & 0	    & 38.04         & 38.73         & 38.30         & 24.37         & 25.45         & 25.45         & 40.36         & 38.55         & 36.65         & 34.26         & 34.25         & 33.47             \\ 

\noalign{\vskip 1mm} 

fog simulation in \cite{STF}        & *     & 41.07         & 41.63         & 40.81         & 21.12         & 21.12         & 21.12         & 38.83         & 37.57         & 34.94         & 33.67         & 33.44         & 32.29             \\ 
our fog simulation                  & *     &\textbf{42.16} &\textbf{42.75} &\textbf{41.70} &\textbf{25.13} &\textbf{25.72} &\textbf{26.22} & 39.19         & 38.29         & 36.29         &\textbf{35.49} &\textbf{35.59} &\textbf{34.74}     \\ 

\noalign{\vskip 1mm} \hline \noalign{\vskip 1mm} 

PointPillars~\cite{PP} $\dagger$    & 0	    & 34.30	        & 35.23	        & 35.00	        & 23.05         & 23.26	        & 25.50         & 26.43	        & 25.35	        & 24.17         & 27.93         & 27.95         & 28.22             \\ 
PointPillars~\cite{PP} $\ddagger$   & 0	    & 34.89         & 35.84         & 35.47         &\textbf{24.14} &\textbf{24.36} &\textbf{25.38} & 27.17         & 26.04         & 24.85         & 28.74         & 28.75         & 28.57             \\ 

\noalign{\vskip 1mm} 

fog simulation in \cite{STF}        & *     & 37.02         & 38.16         & 37.88         & 21.68         & 21.68         & 23.33         & 28.84         & 28.25         & 26.95         & 29.18         & 29.37         & 29.39             \\ 
our fog simulation                  & *     &\textbf{38.31} &\textbf{39.14} &\textbf{38.91} & 23.40         & 23.40         & 25.37         &\textbf{30.50} &\textbf{29.51} &\textbf{27.91} &\textbf{30.73} &\textbf{30.68} &\textbf{30.73}     \\ 

\end{tabular}
\caption{3D average precision (AP) results on the STF~\cite{STF} dense fog test split. \\
$\dagger$ \textit{clear weather baseline} $\ddagger$ \textit{clear weather baseline (same model as $\dagger$) with strongest} $\cap$ \textit{last filter applied at test time} \\ 
* \textit{fog simulation gets applied to every training example with} $\alpha$ \textit{uniformly sampled from} [0, 0.005, 0.01, 0.02, 0.03, 0.06]}
\label{table:3D_classes_relaxed}
\end{table*}
\begin{table*}
\customsize
\begin{tabular}{ cl rrr | rrr | rrr | rrr }
\multirow{2}{*}{Method} & \multirow{2}{*}{$\alpha$} & \multicolumn{3}{c|}{dense fog} & \multicolumn{3}{c|}{light fog} & \multicolumn{3}{c|}{clear}   & \multicolumn{3}{c}{mAP over conditions}                                                                  \\ 

                                    &		& easy		    & mod.		    & hard		    & easy		    & mod.	    	& hard		    & easy		    & mod.		    & hard		    & easy		    & mod.		    & hard              \\ 

\hline \noalign{\vskip 1mm} 

PV-RCNN~\cite{PV-RCNN} $\dagger$    & 0	    & 45.03         & 46.00         & 45.08         & 69.55         & 70.17         & 68.44         & 79.61         & 77.05         & 71.03         & 64.73         & 64.41         & 61.52             \\ 
PV-RCNN~\cite{PV-RCNN} $\ddagger$   & 0	    & 45.24         & 46.18         & 45.25         & 69.64         & 70.30         & 68.42         &\textbf{79.80} & 77.16         & 71.08         & 64.89         & 64.55         & 61.58             \\ 

\noalign{\vskip 1mm} 
fog simulation in~\cite{STF}        & *     & 45.60	        & 46.60	        & 45.60	        & 70.02	        & 70.56	        &\textbf{69.37}	& 79.63	        &\textbf{77.48}	&\textbf{72.48} & 65.08         & 64.88         &\textbf{62.48}     \\ 
our fog simulation                  & *     &\textbf{46.69}	&\textbf{47.38}	&\textbf{46.51}	&\textbf{71.42}	&\textbf{70.96}	& 69.03         & 79.27	        & 76.75	        & 71.80	        &\textbf{65.79} &\textbf{65.03} & 62.45             \\ 

\noalign{\vskip 1mm} \hline \noalign{\vskip 1mm} 

PointRCNN~\cite{PRCNN} $\dagger$    & 0	    & 44.00	        & 45.03	        & 43.73	        & 71.30 	    &\textbf{71.48} &\textbf{68.31} & 80.05         & 76.52         &\textbf{70.80} & 65.12         & 64.34         & 60.95             \\ 
PointRCNN~\cite{PRCNN} $\ddagger$   & 0	    & 44.40         & 45.25         & 44.15         &\textbf{71.36} & 70.45         & 68.28         & 79.96         & 76.37         & 70.59         & 65.24         & 64.02         & 61.01             \\ 

\noalign{\vskip 1mm} 

fog simulation in~\cite{STF}        & *     & 46.08         & 47.02         & 45.85         & 70.80         & 70.27         & 67.66         & 79.90         & 76.16         & 69.18         & 65.59         & 64.48         & 60.90             \\ 
our fog simulation                  & *     &\textbf{47.81} &\textbf{47.99} &\textbf{46.68} & 70.74         & 70.84         & 67.65         &\textbf{80.41} &\textbf{76.58} & 69.68         &\textbf{66.32} &\textbf{65.14} &\textbf{61.34}     \\ 
                                    
\noalign{\vskip 1mm} \hline \noalign{\vskip 1mm} 

SECOND~\cite{SECOND} $\dagger$      & 0	    & 42.36	        & 42.99	        & 41.99	        &\textbf{70.51} & 70.07	        & 68.60	        & 78.67	        & 75.20	        & 70.67	        & 63.85         & 62.75         & 60.42             \\ 
SECOND~\cite{SECOND} $\ddagger$     & 0	    & 42.78         & 43.47         & 42.42         & 70.50         &\textbf{70.08} &\textbf{68.63} & 78.53         & 75.08         & 69.91         & 63.93         & 62.87         & 60.32             \\ 

\noalign{\vskip 1mm} 

fog simulation in~\cite{STF}        & *     & 42.67         & 43.58         & 42.77         & 69.67         & 69.89         & 68.45         & 79.23         &\textbf{76.61} & 71.89         & 63.85         &\textbf{63.36} & 61.04             \\ 
our fog simulation                  & *     &\textbf{43.47} &\textbf{44.01} &\textbf{43.20} & 69.55         & 69.63         & 68.49         &\textbf{79.44} & 75.95         &\textbf{71.94} &\textbf{64.15} & 63.20         &\textbf{61.21}     \\ 

\noalign{\vskip 1mm} \hline \noalign{\vskip 1mm} 

Part-A²~\cite{PartA2} $\dagger$     & 0	    & 37.60	        & 38.15	        & 37.76	        & 65.29	        & 65.88	        & 64.31	        & 76.38	        & 73.79	        & 68.56         & 59.76         & 59.27         & 56.88             \\ 
Part-A²~\cite{PartA2} $\ddagger$    & 0	    & 38.04         & 38.73         & 38.30         & 65.98         & 66.41         & 64.62         & 76.43         &\textbf{73.86} &\textbf{68.57} & 60.15         & 59.67         & 57.16             \\ 

\noalign{\vskip 1mm} 

fog simulation in~\cite{STF}        & *     & 41.07         & 41.63         & 40.81         & 65.91         & 65.84         & 63.91         & 76.61         & 73.84         & 68.31         & 61.20         & 60.44         & 57.68             \\ 
our fog simulation                  & *     &\textbf{42.16} &\textbf{42.75} &\textbf{41.70} &\textbf{68.12} &\textbf{67.76} &\textbf{65.19} &\textbf{76.64} &\textbf{73.86} & 68.05         &\textbf{62.31} &\textbf{61.46} &\textbf{58.32}     \\ 

\noalign{\vskip 1mm} \hline \noalign{\vskip 1mm} 

PointPillars~\cite{PP} $\dagger$    & 0	    & 34.30	        & 35.23	        & 35.00	        & 67.92	        & 68.47	        & 66.73	        & 77.20	        & 74.64	        &\textbf{69.63} & 59.81         & 59.45         & 57.12             \\ 
PointPillars~\cite{PP} $\ddagger$   & 0	    & 34.89         & 35.84         & 35.47         & 67.97         & 68.52         & 66.76         & 77.27         &\textbf{74.66} & 69.59         & 60.04         & 59.67         & 57.27             \\ 

\noalign{\vskip 1mm} 

fog simulation in~\cite{STF}        & *     & 37.02         & 38.16         & 37.88         & 68.18         & 68.24         & 67.10         & 76.33         & 73.91         & 69.03         & 60.51         & 60.10         & 58.00             \\ 
our fog simulation                  & *     &\textbf{38.31} &\textbf{39.14} &\textbf{38.91} &\textbf{68.31} &\textbf{68.95} &\textbf{67.18} &\textbf{77.42} & 74.56         & 69.55         &\textbf{61.34} &\textbf{60.88} &\textbf{58.55}     \\ 

\end{tabular}
\caption{Car 3D AP@.5IoU results on all relevant STF~\cite{STF} test splits. \\
$\dagger$ \textit{clear weather baseline} $\ddagger$ \textit{clear weather baseline (same model as $\dagger$) with strongest} $\cap$ \textit{last filter applied at test time} \\ 
* \textit{fog simulation gets applied to every training example with} $\alpha$ \textit{uniformly sampled from} [0, 0.005, 0.01, 0.02, 0.03, 0.06]}
\label{table:3D_conditions_relaxed}
\end{table*}

\begin{figure*}
     \centering
     \begin{subfigure}[b]{0.325\linewidth}
         \centering
         \includegraphics[width=\linewidth]{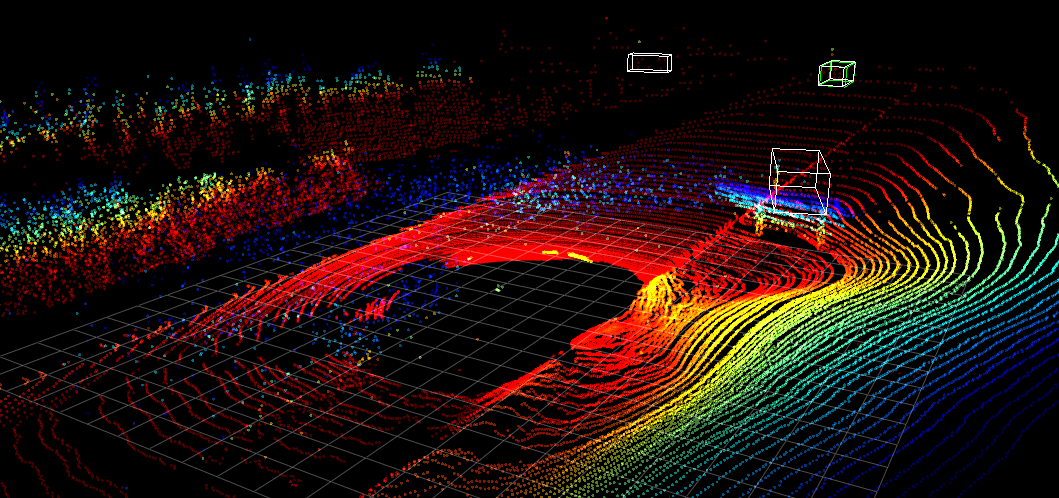}
     \end{subfigure}
     \hfill
     \begin{subfigure}[b]{0.325\linewidth}
         \centering
         \includegraphics[width=\linewidth]{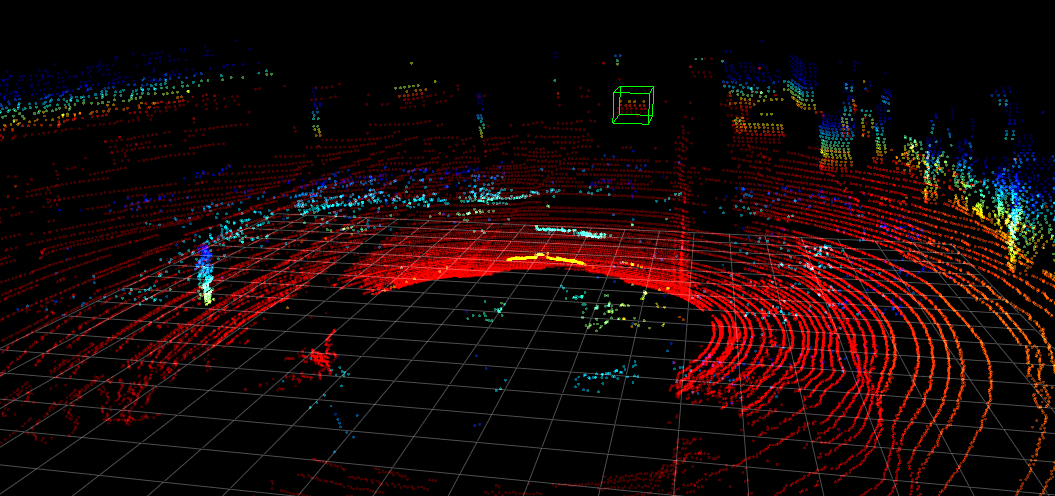}
     \end{subfigure}
     \hfill
     \begin{subfigure}[b]{0.325\linewidth}
         \centering
         \includegraphics[width=\linewidth]{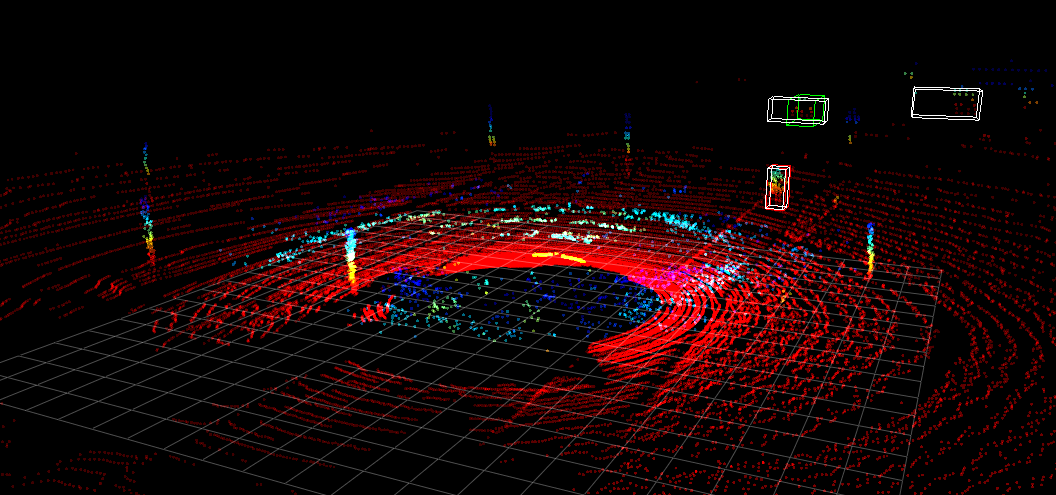}
     \end{subfigure}
     \hfill
     \begin{subfigure}[b]{0.325\linewidth}
         \centering
         \includegraphics[width=\linewidth]{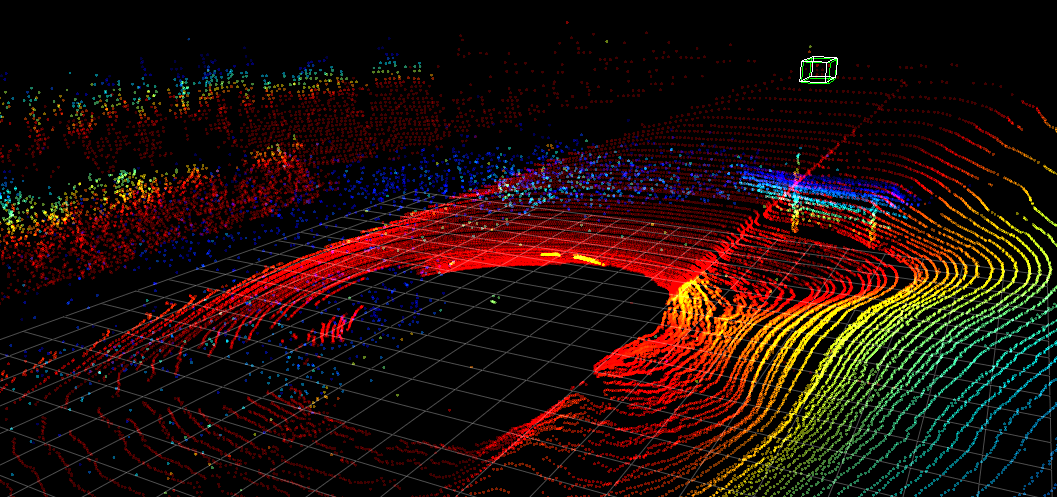}
     \end{subfigure}
     \hfill
     \begin{subfigure}[b]{0.325\linewidth}
         \centering
         \includegraphics[width=\linewidth]{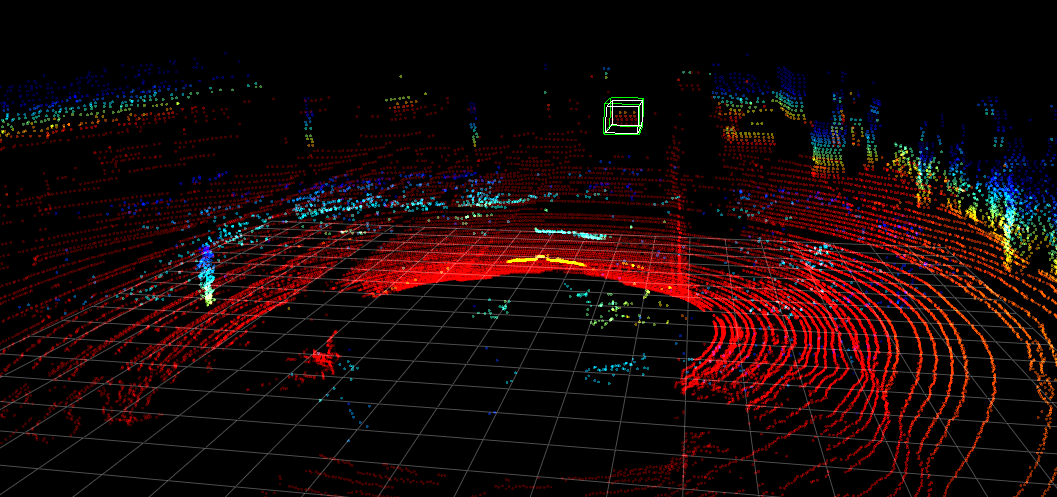}
     \end{subfigure}
     \hfill
     \begin{subfigure}[b]{0.325\linewidth}
         \centering
         \includegraphics[width=\linewidth]{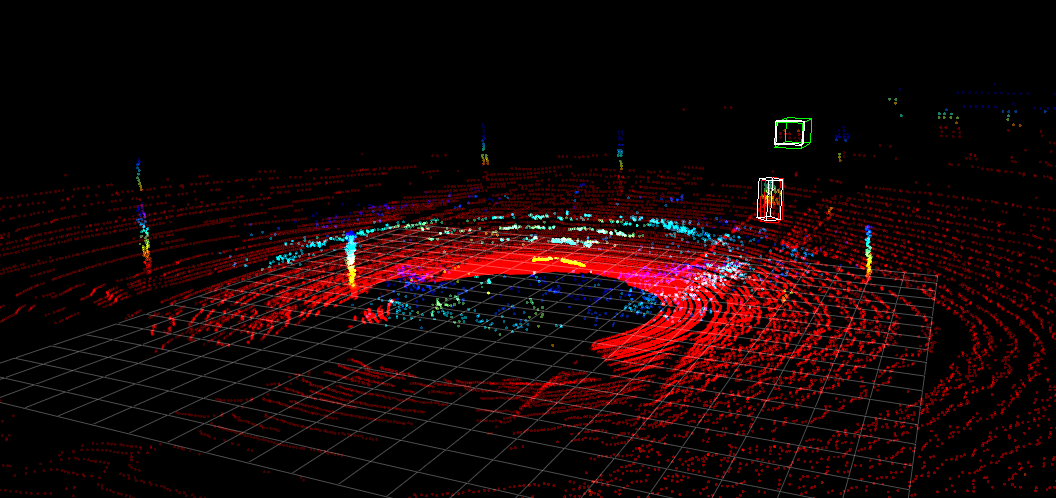}
     \end{subfigure}
     \caption{The (top) row shows predictions by PV-RCNN~\cite{PV-RCNN} trained on the original clear weather data (first row in tables above), the (bottom) row shows predictions by PV-RCNN~\cite{PV-RCNN} trained on a mix of clear weather and simulated foggy data (fourth row in tables above) on three example scenes from the STF~\cite{STF} dense fog test split. Ground truth boxes in color, predictions of the model in white. Best viewed on a screen (and zoomed in).}
     \label{fig:prediction_comparison}
\end{figure*}

\subsubsection{Quantitative Results}
\label{sec:quantitative}

For the numbers we report, we select the snapshot with the best performance on the clear weather validation set and test it on the aforementioned test splits. In Table~\ref{table:3D_classes_relaxed} we report the 3D average precision (AP) on the STF~\cite{STF} dense fog test split for the classes \textit{Car}, \textit{Cyclist} and \textit{Pedestrian} as well as the 3D mean average precision (mAP) over those three classes. Note, it is always one model that predicts all three classes and \textit{not} one model per class. All AP and mAP numbers reported in this paper are being calculated using 40 recall positions as suggested in~\cite{Simonelli_2019_ICCV}. We can see that in mAP over all classes and on the major \textit{Car} class, the training runs of all methods using our fog simulation outperforms the clear weather baseline and the training runs using the fog simulation in~\cite{STF}. 

As a second baseline, we evaluate the clear weather model after applying an additional preprocessing step at test time, where we only feed those points to the network that are present in both, the \textit{strongest} and \textit{last} measurement of the same scene. We dub this filter ``\textit{strongest} $\cap$ \textit{last filter}''. The idea for this filter stems from the fact, that all the points that get discarded by this filter, must be noise (most likely introduced by fog in the scene) and can not be from a physical object of interest. We can see that this filter most of the time boosts the performance of the clear weather model but also does not surpass any fog simulation runs for the majority of cases. One might also notice, that the performance on the \textit{Cyclist} class is generally lower than on the other two classes.  We attribute this to the fact, that the \textit{Cyclist} class is fairly underrepresented compared to the other two classes in the STF~\cite{STF} dataset (e.g.\ $28$ cyclists vs.\ $490$ pedestrians and $1186$ cars in the dense fog test split). For the \textit{Pedestrian} (and \textit{Cyclist}) class we still achieve three out of five times state-of-the-art performance. For the training runs using either our fog simulation or the fog simulation in~\cite{STF}, we uniformly sample for each training example $\alpha$ from [0, 0.005, 0.01, 0.02, 0.03, 0.06] which corresponds to a MOR of approximately [$\infty, 600, 300, 150, 100, 50$]m respectively. Trying out more sophisticated techniques like curriculum learning~\cite{Curriculum} is kept for future work.

In Table~\ref{table:3D_conditions_relaxed} we present the 3D AP of the major \textit{Car} class on the dense fog, light fog and clear test set as well as the mAP over those three weather conditions. We can see that in dense fog the training runs of all methods using our fog simulation outperforms all other training runs, which is exactly what we aimed for with our physically accurate fog simulation. We can further see that mixing in our fog simulation in training does not hurt the performance in clear weather too much, hence we also achieve state-of-the-art for most cases in mAP over all three weather conditions. 

In the Supplementary Materials, we discuss why we chose to focus on relaxed intersection over union (IoU) thresholds and present results using the official KITTI~\cite{KITTI} evaluation strictness. Additionally, we present 2D and Birds Eye View (BEV) results, and further details on the STF~\cite{STF} dataset. 

\subsubsection{Qualitative Results}
\label{sec:qualitative}

In Fig.~\ref{fig:prediction_comparison} we showcase three examples where we clearly outperform the clear weather baseline. We can examine that the model that sees our simulated fog in training, has less false positives (left), more true positives (middle) and overall more accurate predictions (right), each time applying the same confidence threshold for a fair comparison. 
\section{Conclusion}
\label{sec:conclusion}

In this work we introduce a physically accurate way to convert real-world clear weather point clouds into \textit{foggy} point clouds. In this process we have full control over all parameters involved in the physical equations. This not only allows us to realistically simulate any density of fog, but also allows us to simulate the influence of fog on basically any LiDAR sensor currently available on the market.   

We show that by using this physically accurate fog simulation, we can improve the performance of several state of the art 3D object detection methods on point clouds that have been collected in real-world dense fog. We expect that our fog simulation can lead to even greater performance boosts if the LiDAR data is annotated in 360° and not just in the field of view of a single forward facing camera, but no such dataset is publicly available yet to test this hypothesis.

We believe that our physically accurate fog simulation is not just applicable to the task of 3D object detection. So we hope that our fog simulation also finds its way into many other tasks and works. 

\vspace{0.1cm}
\noindent
\textbf{Acknowledgements:} 
This work was funded by Toyota Motor Europe via the research project TRACE Zurich.


{\small
\bibliographystyle{ieee_fullname}
\bibliography{main}
}

\end{document}